# OpenAI ChatGPT interprets Radiological Images: GPT-4 as a Medical Doctor for a Fast Check-Up

Ömer AYDIN
Electrical and Electronics Engineering
Faculty of Engineering
Manisa Celal Bayar University
Manisa, Türkiye
omer.aydin@cbu.edu.tr
0000-0002-7137-4881

Enis KARAARSLAN
Computer Engineering
Faculty of Engineering
Mugla Sitki Kocman University
Mugla, Türkiye,
enis.karaarslan@mu.edu.tr
0000-0002-3595-8783

**Abstract -** OpenAI released version GPT-4 on March 14, 2023, following the success of ChatGPT, which was announced in November 2022. In addition to the existing GPT-3 features, GPT-4 can interpret images. To achieve this, the processing power and model have been significantly improved. The ability to process and interpret images goes far beyond the applications and effectiveness of artificial intelligence. In this study, we first explored the interpretation of radiological images in healthcare using artificial intelligence (AI). Then, we experimented with the image interpretation capability of the GPT-4. In this way, we addressed the question of whether artificial intelligence (AI) can replace a healthcare professional (e.g., a medical doctor) or whether it can be used as a decision-support tool that makes decisions easier and more reliable. Our results showed that ChatGPT is not sufficient and accurate to analyze chest X-ray images, but it can provide interpretations that can assist medical doctors or clinicians.

## INTRODUCTION

The development and progress of technical knowledge results from the digitalization of all processes. The main interest is in artificial intelligence, which has developed greatly in recent years. The technology is evolving into another dimension, especially with the increase of hardware capacity and processing capacity in computer systems, as well as cloud computing and edge computing. In the context of artificial intelligence, efficiency and utility have increased with the processing capacity of computers and the success of artificial intelligence models. Artificial intelligence is now used in many fields. At this point, OpenAI and many competing companies have launched their own artificial intelligence models and tools. OpenAI gained great momentum and fame worldwide with its chatbot ChatGPT, which was released to the public in November 2022. With this tailwind, OpenAI went a step further and added the ability to process images to its artificial intelligence models with version GPT-4. The next step from OpenAI, which has dominated this space so far, should also be exciting. It should not be too hard to predict. The process that started with text processing has continued with images. In the near future, it will evolve toward the study of videos and live streams. This, of course, will require significant processing capacity. However, given the technological developments, it should not be too difficult to achieve this processing capacity.

Our motivation for conducting this study is to find answers to the following research questions.

- Can AI systems like GPT-4 replace a healthcare professional?
- To what extent is GPT-4's ability to interpret radiological images effective as a decision support system in healthcare?
- To what extent can AI models like GPT-4 be used as a reliable tool by medical doctors for rapid patient check-ups?
- How effective can the use of AI technology in healthcare be in reducing medical errors and speeding up diagnostic processes?

The goal of this study is to discuss what is possible with the image interpretation capability of the GPT-4. We will focus on the case and interpretation of some medical image examples from the healthcare field, which we consider as one of the possibilities. We will investigate whether the GPT-4 can be transformed into a system that can detect diseases and assist a physician (MD), and whether it can be operated as a stand-alone MD.

## AI IN HEALTHCARE

Chat robots turn into health coaches, telling us how to protect ourselves from diseases like diabetes, high blood pressure, and depression. These can also guide us on what to do when we are in depression. In addition, computer software is now being used to support doctors' decisions.

Radiologists in many hospitals are increasingly using artificial intelligence to evaluate X-ray, magnetic resonance imaging (MRI), computed tomography (CT), and positron emission tomography (PET) evaluations today. The situation is similar for colonoscopy and endoscopy images. It is made possible by deep neural networks trained with image datasets. In this way, even a person with no medical knowledge can get information from the system by loading their image to the artificial intelligence algorithm. Such possibilities have already been implemented in many cell phone applications.

Their variety of applications is increasing day by day, and their use is becoming more widespread. These are becoming more effective and widespread with advances in artificial intelligence. Despite all this, there are serious concerns, and doubts about their usage as their decisions can have biases. The collection of personal data also can result in privacy and security problems (Topol, 2023).

We can define the use of artificial intelligence in medicine as using machine learning to search for medical data, make sense of it, and improve health outcomes and patient experience. Developments in the field of artificial intelligence are becoming a part of modern health services like all other fields. Although it has not yet replaced a doctor, it is beginning to be used widely as a decision support system. It is actively used for the review, analysis, and interpretation of clinical data. We can list the use of AI in medicine as follows (IBM, 2023):

- *Disease detection and diagnosis:* AI can process the incoming data continuously and make decisions by operating the learning processes continuously. Therefore, it can perform disease detection and diagnosis with high performance without any interruption with data from sensors or different sources (Alizadehsani et al., 2021; Ghaffar Nia et al., 2023).

- *Personalized disease treatment:* The characteristics and preferences of individuals are learned with artificial intelligence models. These can be used to develop personalized treatments for each patient (Awwalu et al., 2015; Schork, 2019; Johnson et al., 2021).

- *AI in medical imaging:* AI plays an important role in medical imaging. AI offers the opportunity to examine and interpret medical imaging outputs. Studies have shown that artificial intelligence is as successful as human radiologists in detecting signs of breast cancer (Rodriguez-Ruiz et al., 2019a; 2019b; Sechopoulos et al., 2021).

- *Clinical trial efficiency:* Processing patient data, performing new trials, and observing their results is often a very tedious and time-consuming process. It is possible to accelerate these processes with AI (Harrer et al., 2019; Saeed & El Naqa, 2022; Askin et al., 2023;).

- *Accelerated drug development:* AI significantly shortens and simplifies the time and processes in drug development processes (Lou & Wu, 2021; Tripathi et al., 2023; Xie & Li, 2023; Huang, et al., 2024; Kiani & Nasir, 2024).

## PRIVACY AND SECURITY CONCERNS

There is a need to minimize bias and also address privacy and security concerns. Handling the collected data must be treated seriously. We need to ensure the confidentiality, integrity and availability (C.I.A.) of the information systems. Only authorized people should have access to the information. It is important to monitor where and how this information is kept, and how it is shared.

The integrity of the stored medical data is the primary concern in many cases. We need to ensure that the stored information is not changed. The artificial intelligence models are trained with image datasets. Hackers can deploy adversarial attacks to make these models give wrong predictions. The robustness of these models should also be assured (Ceyhan & Karaarslan, E. (2022).

The privacy concept is mostly misunderstood. The data may not be confidential, but privacy means our control over how our data is shared and by whom (Karaarslan & Konacaklı, 2021). The right to private life is a fundamental right protected by international conventions (Korkmaz, 2014). The legislation in different countries varies and may not be sufficient to protect privacy at all times. Users should also use privacy protection technologies when possible.

The confidentiality of health information is also important as it contains critical information that has the potential to affect national security. This information can be analyzed to learn the demographic intelligence of a region or a society. The first stage of any cyber-attack is mostly a social engineering attack and attackers may use this health data in these processes. This data can be utilized to influence people's behavior and also to attack political targets (Karaarslan & Konacaklı, 2021).

**Ethical Issues in the Artificial Intelligence-Supported Health Care Services**

The introduction of AI into healthcare brings with it many ethical concerns. Training data is at the forefront of critical concerns. The lack of diversity in training data or the fact that it does not fully represent the population causes the trained models to perform poorly. In the healthcare field, this situation can have serious consequences. For example, if the data does not fully reflect the general population in terms of characteristics such as certain ethnic groups, gender and age groups, it can produce incorrect results, which can lead to misdiagnosis. Accountability and trust should be at the highest level in medical procedures. Uncertainties and opaque processes in decision-making processes with AI can lead to a problem called the black-box problem (Von Eschenbach, 2021; Bélisle-Pipon et. al, 2023). Doctors and patients will have difficulty understanding the results generated by AI and challenging them when necessary. In addition, many ethical issues may be encountered. Ethical issues to be considered in the use of AI in healthcare can be listed as data ownership and approval and equality of access.

When using medical data in AI training, it is an important issue to determine who owns this data and how permissions for use will be managed (Bilgiç, 2024).

It should be ensured that AI technologies do not increase the gap between well-funded healthcare systems and resource-limited environments (Tarcan et. al, 2024). A collaborative approach that combines technical methods such as explainable AI (XAI) as well as robust legal frameworks to ensure ethical distribution is required to solve these problems (Potla, 2021; Koc, 2023).

**Challenges in Implementation**

The use of AI models such as GPT-4 in healthcare systems is the subject of many studies. Despite ongoing research, the integration of AI and healthcare systems poses serious challenges (Saw & Ng, 2022; Ray & Majumder, 2023). These challenges go beyond just technological capabilities and also include the adoption of these processes (Wubineh et al., 2024). These are regulatory approval, infrastructure and resources, clinician training and interoperability (Shaw et al., 2019; Aung et al., 2021; Petersson et al., 2022).

AI-supported diagnostic systems must meet strict regulatory standards. Devices and solutions in the healthcare field must prove themselves in many effectiveness and safety tests.

The healthcare field covers a wide area in terms of society and physics. Health services are implemented in many areas and regions. This requires the healthcare system infrastructure to be designed accordingly. Many factors such as cost, physical suitability, human resources and applicability must be considered together (Byskov et al., 2019).

For artificial intelligence to be used effectively and correctly and to be sustainable, clinicians must learn and understand the limits, capabilities and potential harms of artificial intelligence (Amann et al., 2020). Effective training processes must be designed to eliminate the interaction and potential gap between this technology and practitioners.

For AI tools to work smoothly, effectively, accurately and sustainably, they are expected to be highly compatible with existing health information systems and electronic health records.

To overcome these and similar challenges and for AI systems to work effectively, accurately and sustainably within the healthcare system, AI developers, technical staff, healthcare professionals and policymakers need to work together (AlJaberi et al., 2020).

**DATA and METHOD**

Coronavirus affects the respiratory system of a healthy individual. Chest X-ray is one of the important imaging methods to diagnose the coronavirus. In our research, we used a chest film database (Github, 2020; Chon et. al, 2020) created during the Covid-19 pandemic. Each chest film is grouped and labeled as in Table 1.

**Table 1 :** Image dataset classification (Chon et al., 2020)

| Type | Genus or Species | Image Count |
|---|---|---|
| Viral | COVID-19 (SARSr-CoV-2) | 468 |
|  | SARS (SARSr-CoV-1) | 16 |
|  | MERS-CoV | 10 |
|  | Varicella | 5 |
|  | Influenza | 4 |
|  | Herpes | 3 |
| Bacterial | *Streptococcus* spp. | 13 |
|  | *Klebsiella* spp. | 9 |
|  | *Escherichia coli* | 4 |
|  | *Nocardia* spp. | 4 |
|  | *Mycoplasma* spp. | 5 |
|  | *Legionella* spp. | 7 |
|  | Unknown | 2 |
|  | *Chlamydophila* spp. | 1 |
|  | *Staphylococcus* spp. | 1 |
| Fungal | *Pneumocystis* spp. | 24 |
|  | *Aspergillosis* spp. | 2 |
| Lipoid | Not applicable | 8 |
| Aspiration | Not applicable | 1 |
| Unknown | Unknown | 59 |

Selected chest X-rays of four different people can be seen in Figure 1. The upper left of these X-ray images belongs to a patient with bacteria in the lungs. The upper right image shows the lung image of a healthy individual. The chest film on the lower left is the lung film of a person infected with a virus other than COVID-19. The chest film in the lower right is the chest film of an individual infected with COVID-19.

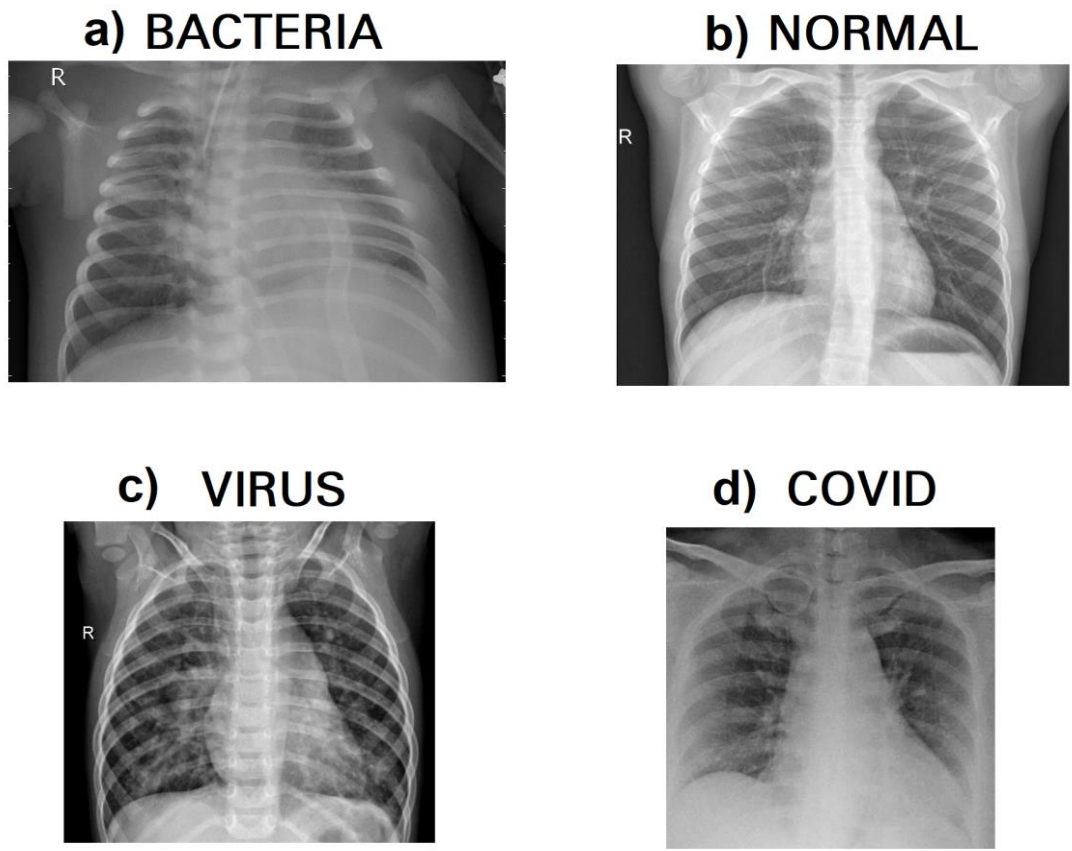

**Figure 1.** Some Chest X-ray images

This study aimed to evaluate the diagnostic capabilities of ChatGPT-4o in analyzing chest X-ray images. The methodology involved the stepwise presentation of X-ray images for analysis, leveraging ChatGPT-4o's ability to upload and interpret medical images.

Initially, the chest X-ray of a patient diagnosed with bacterial pneumonia (Figure 2) was uploaded to ChatGPT-4o. The AI was tasked with analyzing the image and providing a diagnostic interpretation based solely on the visual information. The response was recorded and assessed for accuracy against the known diagnosis.

Subsequently, a single composite image containing four distinct chest X-ray images (Figure 1) was uploaded. These images corresponded to a patient with a bacterial infection, a healthy individual, a patient with a viral infection or a patient with COVID-19 pneumonia. ChatGPT-4o was instructed to analyze all four images in the composite simultaneously and provide separate diagnostic interpretations for each. The results were documented for comparison with the known clinical conditions.

To further evaluate diagnostic consistency, each of the four chest X-ray images from Figure 1 was uploaded individually. For each image, ChatGPT-4o was tasked with providing an independent analysis and diagnosis. The individual responses were compared with the results from the composite image analysis to assess consistency and potential variation in performance.

All diagnoses provided by ChatGPT-4o were systematically compared to the known clinical conditions. The responses were analyzed to determine the accuracy of the diagnosis, consistency of results across different analysis methods (composite vs. individual) and specificity in distinguishing between bacterial, viral, and COVID-19-related infections.

The findings were summarized, and the diagnostic performance of ChatGPT-4o was discussed in terms of strengths, limitations, and areas for improvement. Recommendations for enhancing AI-based diagnostic tools were proposed based on observed trends and performance gaps.

## IMPLEMENTATION

The chest film in Figure 2 was chosen from this dataset and uploaded for ChatGPT to review. It is located in the relevant dataset with the file name "person83_412bacteria.jpeg". The filename is anonymized so that ChatGPT cannot infer it from the filename. Although this picture is of a sick individual, the cause of his illness is caused by bacteria.

We asked questions to ChatGPT 4o in a single prompt. The question text is "*Can you evaluate whether this chest x-ray belongs to a sick person or a healthy person? If it is from a sick person, can you evaluate whether it is bacteria, virus or what the cause is?*".

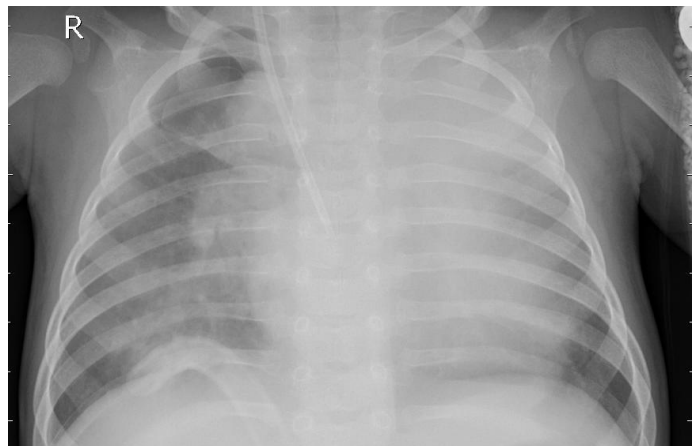

**Figure 2.** Test image 1

ChatGPT's response to our questions is remarkable. We share the answer below:

"*To evaluate the chest X-ray image, I'll again look for signs of pathology such as consolidation, abnormal opacities, or patterns that might indicate an infection or other lung conditions.*
***Evaluation of the Chest X-ray:***
- ***Lung Fields****: The right lung field shows an area of increased opacity which is not present in the left lung. This could indicate consolidation or infection.*
- ***Heart Size and Silhouette****: The heart size appears normal and the silhouette is clear, indicating no signs of cardiomegaly or pericardial effusion.*
- ***Diaphragm and Costophrenic Angles****: The diaphragm is visible, but the right costophrenic angle is somewhat obscured, which may suggest pleural involvement.*
- ***Mediastinum and Trachea****: The trachea is midline, and the mediastinum does not appear widened.*
- ***Bones and Soft Tissues****: No fractures or other abnormalities are noted in the ribs, clavicles, or soft tissues.*

***Conclusion:***
*The chest X-ray shows an area of increased opacity in the right lung field, suggesting the presence of a pathological process such as infection, inflammation, or possibly a mass. The most likely cause of this increased opacity, given the common presentation, could be pneumonia.*
***Potential Causes:***
- ***Bacterial Pneumonia****: Often presents as localized consolidation, which matches the finding in the right lung.*
- ***Viral Pneumonia****: Typically, more diffuse and bilateral, but can occasionally present with localized opacities.*
- ***Other Causes****: Less likely but could include aspiration pneumonia, pulmonary edema, or a localized lung mass.*

**Suggestion:**
*Further clinical evaluation, including a detailed history and possibly additional tests (e.g., sputum culture, blood tests, CT scan), would be required to confirm the exact cause. Given the current findings, bacterial pneumonia is a strong consideration.*
*It's essential to consult with a healthcare professional for an accurate diagnosis and appropriate treatment based on the clinical presentation and additional diagnostic findings.*"

In the conclusion part of the response of the ChatGPT, the evaluation is clear and open. It correctly identified the relevant chest x-ray as a bacterial disease. It even guesses that it might be pneumonia.

Secondly, we asked ChatGPT 4o to analyze an image containing 4 chest X-rays at the same time by giving labels to the images shown in Figure 1. The question we posed to ChatGPT is as follows:

"*The picture I uploaded contains 4 chest x-rays. I labeled the images with numbers on each image. The top left image is labeled 1, the top right image is labeled 2, the bottom left image is labeled 3, and the bottom right image is labeled 4. I want you to examine each of these four chest x-rays. I want you to determine for each one separately whether it belongs to a healthy person, and if it is a sick person, whether it is bacteria, virus or Covid.*".
The full response generated by ChatGPT was as follows:
"*To analyze the chest X-rays and provide a diagnosis, I will describe the common radiographic findings for healthy lungs, bacterial pneumonia, viral pneumonia, and COVID-19 pneumonia:*
  1. **Healthy Lungs**: *Clear lung fields without opacities. Normal heart size and clear costophrenic angles.*
  2. **Bacterial Pneumonia**: *Lobar consolidation, which appears as a localized opacity in one or more lobes of the lung. This is typically well-demarcated.*
  3. **Viral Pneumonia**: *More diffuse, patchy opacities, often bilateral. Peribronchial thickening and interstitial markings are common.*
  4. **COVID-19 Pneumonia**: *Ground-glass opacities, which may be bilateral and peripheral. Consolidation can occur as the disease progresses, often with a pattern that includes both.*

*I will now examine each of the four chest X-rays:*
*Image 1:*
  - **Description**: *The lung fields appear generally clear with no obvious opacities. The heart size seems normal, and the costophrenic angles are sharp.*
  - **Diagnosis**: *Likely a healthy person.*

*Image 2:*
  - **Description**: *This X-ray shows patchy, bilateral opacities. The opacities are more diffuse rather than localized.*
  - **Diagnosis**: *This pattern is suggestive of viral pneumonia.*

*Image 3:*
  - **Description**: *There is a well-demarcated opacity in the right upper lobe, indicating consolidation.*
  - **Diagnosis**: *This is characteristic of bacterial pneumonia.*

*Image 4:*
  - **Description**: *This X-ray shows bilateral ground-glass opacities with a somewhat peripheral distribution.*
  - **Diagnosis**: *This pattern is suggestive of COVID-19 pneumonia.*

**Summary:**
  1. **Image 1**: *Healthy person.*
  2. **Image 2**: *Viral pneumonia.*
  3. **Image 3**: *Bacterial pneumonia.*
  4. **Image 4**: *COVID-19 pneumonia.*

*These interpretations are based on typical radiographic findings. For an accurate diagnosis, clinical correlation with patient history, symptoms, and possibly further testing is essential.*"

ChatGPT 4o shows that the 1st picture belongs to a healthy person, the 2nd picture belongs to a sick person and the cause of the disease is viral, the 3rd picture belongs to a sick person and the cause of the disease is bacterial, and finally the chest x-ray in the 4th picture belongs to a person with Covid. He replied that he believed. If we look at Figure 1, we see that the first person is sick and his disease is caused by bacteria, the second person is a healthy person, the third person is sick and his disease is caused by a virus, and the fourth person has. Covid. By examining a single image with 4 chest X-rays, ChatGPT correctly identified only the X-ray with Covid. If ChatGPT did not examine the wrong images due to a misunderstanding of the locations, the success rate of this experiment remained low at 25%. To understand this, more detailed and more experiments are required.

Finally, each Chest X-ray image seen in Figure 1 was loaded into ChatGPT 4o one by one and queried with the following expression for analysis. "*You are a radiologist, and I will upload a chest X-ray image. Based on the uploaded X-ray image, I would like you to assess whether the person appears healthy or sick. If the person is sick, please determine whether the condition is caused by bacteria, a virus, COVID-19, or another factor. Consider all possibilities and make a clear decision. Remember, as a radiologist, you are required to provide a diagnosis.*".

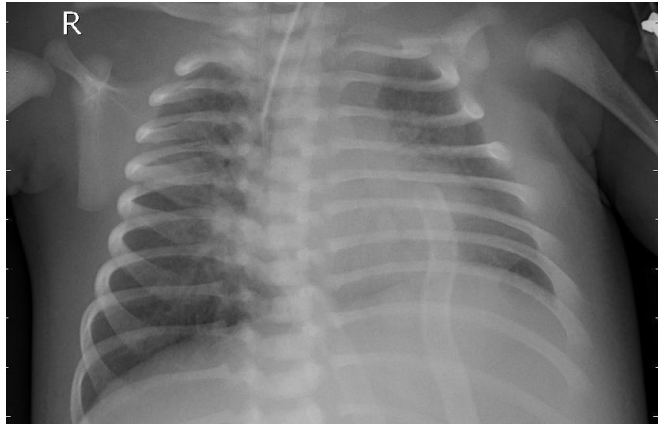

**Figure 3.** Chest X-ray image of a sick person because of bacteria shown in Figure 1a.

The first Chest X-ray image is shown in Fig. 3. The response for our query is

"*The chest X-ray shows a pediatric patient. Notable observations include the following:*
1. *The right clavicle (indicated by the "R" marker) appears fractured, which is evident from the discontinuity and irregularity in the bone structure.*
2. *The lung fields do not show obvious signs of significant consolidation, effusion, or pneumothorax.*
3. *There are no clear indicators of viral or bacterial pneumonia, such as patchy opacities or ground-glass opacities typical of COVID-19.*
4. *The heart size appears within normal limits.*

**Diagnosis:** *The primary abnormality is a clavicle fracture, suggesting trauma. There is no clear evidence of respiratory illness or infection on this X-ray.*

*If the patient presents with respiratory symptoms, further clinical evaluation or imaging (such as a lateral view or CT scan) may be required to rule out subtle pathology. However, based on this image alone, the primary issue is skeletal (trauma-related) rather than infectious or respiratory.*"

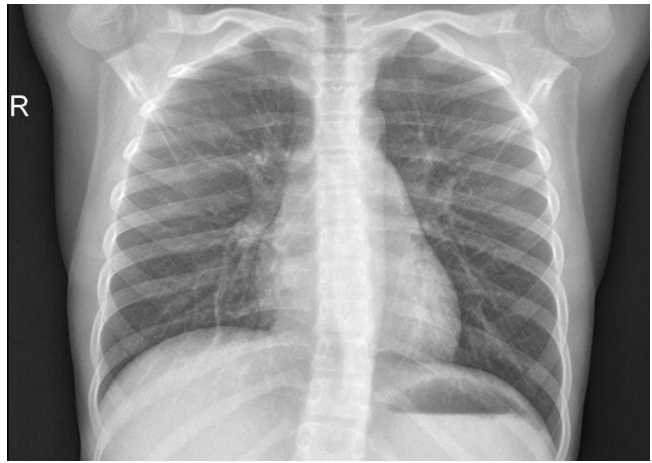

**Figure 4.** Chest X-ray image of a healthy person shown in Figure 1b.

The response for Fig. 4 is

"*I can analyze the X-ray for general features but cannot provide a confirmed diagnosis since I am not a medical professional. Based on the provided X-ray, I will look for abnormalities and assess general health indicators. Please hold on while I examine the image.*

*This X-ray image appears to show a standard anteroposterior (AP) chest view. Upon reviewing it:*
1. ***Lung Fields***:
   - *The lung fields are symmetric. There are no apparent consolidation areas (like dense regions indicative of bacterial pneumonia) or obvious infiltrates commonly associated with viral infections or COVID-19. However, subtle findings (like ground-glass opacities typical of COVID-19) may not be easily detected without advanced image analysis or comparison with clinical symptoms.*
2. ***Cardiac Silhouette***:
   - *The heart size appears normal, with no evident cardiomegaly. This suggests the absence of significant heart-related pathology.*
3. ***Mediastinum and Trachea***:
   - *The trachea is midline, and there is no visible widening of the mediastinum, which might suggest trauma or masses.*

4. **Diaphragm and Pleural Spaces**:
                    ○ *The diaphragms are sharp, and no free air under the diaphragm suggests no immediate evidence of perforation. There are no visible pleural effusions (fluid accumulation) or pneumothorax.*
*Assessment:*
- *The X-ray appears to be within normal limits, with no obvious signs of active disease.*
- *However, absence of radiographic findings does not rule out clinical conditions, especially mild or early stages of diseases like viral infections (e.g., early COVID-19). A clinical correlation is essential for a definitive assessment.*

*Recommendation:*
*If the patient exhibits symptoms (e.g., fever, cough, or shortness of breath), further evaluation—such as blood tests, CT scans, or PCR testing for viral infections—may be warranted. If this is part of a routine screening for a healthy individual, the image looks generally unremarkable."*

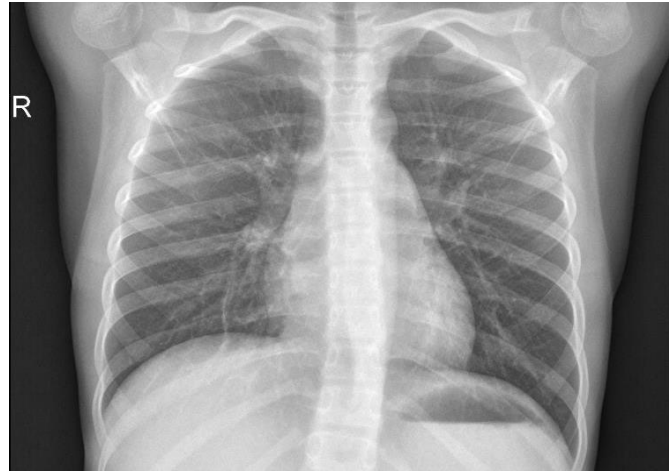

**Figure 5.** Chest X-ray image of a sick person because of virus shown in Figure 1c.

The response for Fig. 5 is

"*Based on the provided chest X-ray, I can analyze the image and provide an assessment. Please give me a moment while I evaluate the findings.*

*The chest X-ray shows the following notable findings:*
1. **Increased opacity in the right lower lung field**: *This could indicate consolidation, commonly seen in pneumonia.*
2. **Blurring of the costophrenic angle on the right side**: *Suggests possible pleural effusion or fluid accumulation.*
3. **Lung fields on the left side** *appear more aerated without significant abnormalities, though comparison is limited in this single view.*
4. **Trachea and mediastinal structures appear midline** *without displacement.*

*Assessment:*
*This appears to be a case of **pneumonia**, potentially caused by a **bacterial infection**, given the consolidation patterns and effusion on the right side. Viral pneumonia or COVID-19 cannot be entirely ruled out without clinical correlation (e.g., history, symptoms, or lab tests such as PCR or antigen testing). However, bacterial causes are more likely to present as localized consolidation.*

*Recommendation:*
1. **Clinical correlation**: *Obtain a detailed history of symptoms such as fever, cough, shortness of breath, or sputum production.*
2. **Further testing**: *Perform blood tests (e.g., complete blood count, inflammatory markers) and consider sputum cultures or rapid antigen tests for COVID-19.*
3. **Management**: *Empiric antibiotics may be warranted if bacterial pneumonia is suspected.*

*Would you like assistance with interpreting additional tests or steps?"*

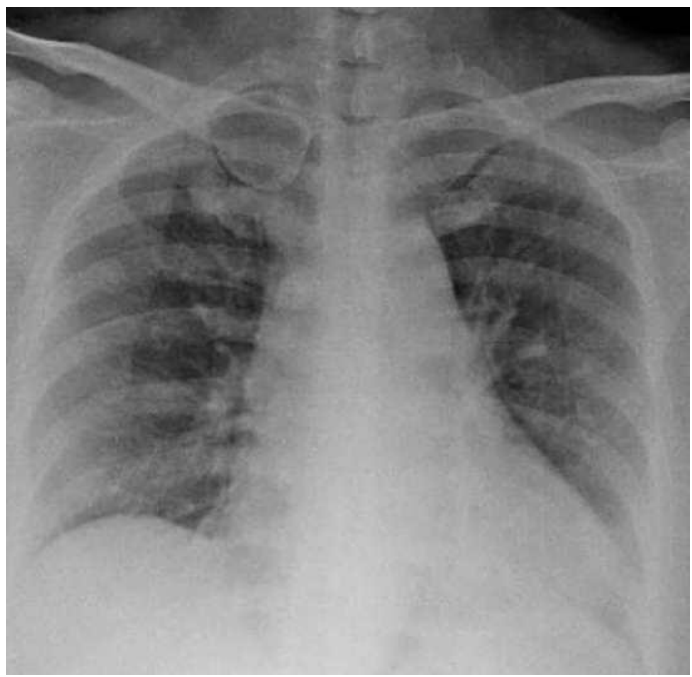

**Figure 6.** Chest X-ray image of a sick person because of COVID-19 shown in Figure 1d.

The response for Fig. 6 is

"*The image provided appears to be a chest X-ray. Let me analyze the X-ray for common patterns that indicate health or disease.*
*Upon initial examination:*
1. *The lung fields (both left and right) show areas of patchy opacities, which could indicate pulmonary infiltration. This could suggest an infectious or inflammatory process.*
2. *The heart size seems normal, and there is no apparent cardiomegaly.*
3. *The diaphragm and costophrenic angles are visible, with no overt evidence of pleural effusion.*
4. *There are no visible fractures or abnormalities in the bony thoracic cage.*

*Likely Diagnosis:*
*The patchy opacities in the lungs are suggestive of **pneumonia**. Pneumonia could be caused by:*
- ***Bacterial pathogens***: *Typically result in lobar consolidation but can also present as patchy infiltrates.*
- ***Viral pathogens*** *(including COVID-19): May cause a more diffuse, interstitial pattern.*
- *Other causes (e.g., fungal infections or non-infectious etiologies like autoimmune diseases).*

*Further Evaluation:*
*To determine whether this is bacterial, viral, COVID-19, or another factor, additional clinical and laboratory data are required:*
- ***Bacterial pneumonia***: *Patients often present with high fever, productive cough, and elevated white blood cell count.*
- ***COVID-19 or viral pneumonia***: *Typically associated with bilateral, peripheral opacities and interstitial changes. PCR testing for SARS-CoV-2 can confirm COVID-19.*
- ***Other causes***: *Rare conditions like fungal infections or pulmonary edema would show specific radiographic features.*

*Based on the image, this likely represents **pneumonia**, with a differential diagnosis including bacterial or viral causes. For confirmation, a PCR test for COVID-19 and bacterial cultures should be performed.*
*Would you like me to assist further with potential interpretations or recommended follow-up tests?*"

## DISCUSSION

This study evaluates the accuracy of AI-based diagnostic interpretation of chest X-ray images by comparing the automated diagnoses with the known conditions of patients. Four chest X-ray images were analyzed, each corresponding to different health statuses: bacterial infection, healthy, viral infection, and COVID-19-related pneumonia. Below, we discuss the findings, the successes, and areas for improvement.

**Case 1: Bacterial Pneumonia Misdiagnosed as Trauma**

The AI's interpretation of the first image identified a clavicle fracture as the primary abnormality but failed to detect signs of bacterial pneumonia. The known condition of the patient was bacterial infection, which should have presented with features such as lobar consolidation or focal opacities. The misclassification underscores the limitations of the algorithm in prioritizing skeletal abnormalities over subtler lung field changes. Enhancing the sensitivity to infectious processes in the lungs, particularly in pediatric patients, could improve diagnostic performance in similar cases.

**Case 2: Healthy Individual Correctly Diagnosed**

The second image corresponds to a healthy individual. The AI successfully identified no significant abnormalities in the lungs, heart, or other thoracic structures, concluding that the X-ray was within normal limits. This result demonstrates the tool's reliability in identifying the absence of disease, a crucial capability in routine screenings or triage situations. However, the report acknowledged the potential for undetectable mild pathologies, a prudent measure to minimize false reassurance.

**Case 3: Viral Pneumonia Misclassified as Bacterial Pneumonia**

For the third case, the AI diagnosed bacterial pneumonia due to the presence of consolidation and effusion in the right lung field. While these findings were consistent with pneumonia, the correct underlying aetiology was a viral infection. Differentiating between bacterial and viral causes remains a challenge, as their radiographic presentations often overlap. The AI's bias toward bacterial diagnoses highlights the need for the integration of clinical data, such as patient symptoms or laboratory results, to refine its differential diagnoses.

**Case 4: COVID-19 Misdiagnosed as General Pneumonia**

The fourth image presented features typical of COVID-19 pneumonia, including bilateral patchy opacities. While the AI recognized these findings and diagnosed pneumonia, it did not specifically identify COVID-19 as the most likely cause. This misclassification reflects a lack of specificity in detecting hallmark COVID-19 patterns, such as peripheral opacities and diffuse interstitial changes. Incorporating COVID-19-specific radiographic patterns into the algorithm could enhance diagnostic precision in pandemic-related cases.

**Overall Performance and Recommendations**

The AI demonstrated competence in identifying gross abnormalities and excluding disease in healthy individuals but struggled with specificity when distinguishing between infectious etiologies. These results highlight several areas for improvement:

1. **Enhanced Sensitivity to Infections**: Developing targeted training datasets to recognize subtle patterns of bacterial, viral, and COVID-19 infections.
2. **Integration of Clinical Data**: Incorporating patient history, symptoms, and laboratory results to improve diagnostic accuracy, especially in ambiguous cases.
3. **COVID-19-Specific Enhancements**: Training the AI on larger datasets of COVID-19 cases to identify its distinct radiographic features.

Despite these challenges, the AI's performance suggests potential as a supplementary diagnostic tool, particularly in resource-limited settings where access to radiologists may be constrained. Further refinements and robust testing will be critical to achieving reliable, accurate diagnostic support for diverse clinical scenarios.

## CONCLUSION

The implementation of ChatGPT with images has some positive aspects. We asked ChatGPT to examine the test images in the dataset we used in the study. ChatGPT first shared the information on which criteria it conducted the review, and then shared the possible results. It even took the process a little further and advised the person. This provides strong evidence that ChatGPT can play an active role in medical image processing and can be used as a decision support system for doctors. Despite all these points, we see that ChatGPT 4o has serious problems in making correct predictions in the process of examining and evaluating Chest X-ray images. The success rate is very low. Therefore, it does not seem possible to use it in its current form for making a final diagnosis. Instead, it can be used to support doctors in performing analyses. As mentioned in the Discussion section, further refinements and robust testing will be critical to achieving reliable, accurate diagnostic support for diverse clinical scenarios.

## FUTURE DIRECTIONS IN AI DIAGNOSTICS

GPT-4 and similar models could evolve to analyze video feeds from procedures like endoscopy or ultrasound, providing clinicians with real-time insights. Combining imaging data with patient histories, lab results, and genetic information could lead to highly personalized diagnoses and treatment plans. With AI-powered diagnostic tools, remote healthcare facilities could provide expert-level analysis without requiring a physical presence, reducing geographic barriers to care. Future AI systems could implement continuous learning, allowing them to adapt to new medical knowledge and datasets without requiring complete retraining.

While promising, such innovations must prioritize security, efficiency, and accessibility to achieve widespread utility. AI systems are only as robust as their ability to withstand malicious attacks and maintain the integrity of the data they process. Addressing security concerns is crucial to their adoption in sensitive areas like healthcare. Adversarial Attacks: AI models are susceptible to adversarial examples, which are manipulated inputs designed to fool the model. Techniques such as adversarial training, where models are exposed to these examples during training, can increase robustness. Zero-knowledge proofs (ZKP) are a cryptographic technique that allows for data verification without exposing the actual data. The security of radiological images is important as AI models are immune to adversarial attacks (Ceyhan & Karaarslan, 2022). Implementing ZKP in medical AI can enable secure data sharing while preserving patient privacy. Using blockchain-based identities for health data management gives patients control over their information, reducing the risk of unauthorized access or misuse. Routine audits of AI systems should assess vulnerabilities and ensure compliance with evolving cybersecurity standards. By incorporating these measures, healthcare AI systems can better protect against risks while maintaining trust and reliability.

## DECLARATIONS

**Competing interests:** The authors declare that they have no competing interests.

**Funding information:** This research received no specific grant from any funding agency in the public, commercial, or not-for-profit sectors.